\documentclass[letterpaper, 10 pt, conference]{ieeeconf}  

\IEEEoverridecommandlockouts                              

\usepackage{mathptmx}
\usepackage{amsmath,amssymb,bm}
\usepackage{graphicx}
\usepackage{capt-of}
\usepackage{hyperref}
\usepackage{booktabs,multirow,array}
\usepackage{cite}
\usepackage{balance}
\usepackage[table]{xcolor}
\usepackage{url}
\newcommand{\simthree}{\mathrm{Sim}(3)}
\newcommand{\R}{\mathbb{R}}
\newcommand{\fail}{--}
\newcommand{\best}[1]{\textbf{#1}}
\newcommand{\second}[1]{\underline{#1}}

\definecolor{grouprow}{gray}{0.90}

\newcommand{\ours}{CoMo3R-SLAM}
\definecolor{oursgray}{gray}{0.93}

\title{\LARGE \bf
CoMo3R-SLAM: Collaborative Monocular Dense SLAM with Learned 3D Priors for Outdoor Multi-Robot Systems
}

\author{Zhihao Cao$^{1}$, Qi Shao$^{2,3}$, Shuhao Zhai$^{4,5}$, Feng Tian$^{1}$,
Anh Nguyen$^{3}$, Hesheng Wang$^{6}$, and Baoru Huang$^{3,7}$%
\thanks{$^{1}$ETH Zurich, Switzerland.
        $^{2}$Harbin Engineering University, China.
        $^{3}$University of Liverpool, United Kingdom.
        $^{4}$University of Macau, China.
        $^{5}$University of Ottawa, Canada.
        $^{6}$Shanghai Jiao Tong University, China.
        $^{7}$Imperial College London, United Kingdom.}%
}

\begin{document}

\maketitle
\thispagestyle{empty}
\pagestyle{empty}

\begin{abstract}
Outdoor robot teams need a shared dense map despite limited overlap, independent reference frames, and uncertain monocular scale. Collaborative dense SLAM systems typically resolve this with depth sensors, which add payload, power, and calibration cost. We present CoMo3R-SLAM, a collaborative monocular dense SLAM system that places learned feed-forward 3D reconstruction priors at the center of the multi-agent problem: their dense pointmaps anchor scale across agents and supply correspondences strong enough to verify inter-agent links geometrically. Each agent tracks and fuses its own keyframes from a single RGB stream, while a coordinator retrieves cross-agent keyframes over the prior's encoder features, verifies them by bidirectional dense pointmap matching, synchronizes the independent similarity gauges in closed form, and refines every keyframe in one unified multi-agent sim(3) graph. Finally, a pose-depth alternation over geometry-aware segments lets inter-agent observations constrain dense structure as well as trajectories. Requiring neither measured depth nor supplied intrinsics, CoMo3R-SLAM attains the lowest trajectory error on three of four Tanks and Temples scenes, and competitive accuracy on Waymo driving sequences, while running at approximately 6-8 FPS on RTX 3080 Ti. A long-horizon traversal, independently captured day and night streams, and teams of up to four agents further map its operating range.
\end{abstract}

\section{Introduction}
\label{sec:intro}
Outdoor robots performing environmental monitoring, or search and rescue benefit from sharing observations of the same environment. Collaborative SLAM provides the common spatial representation needed to combine their trajectories and maps~\cite{schmuck2021covins,tian2022kimera,lajoie2020door}. Dense geometry adds surfaces and free-space context beyond the sparse landmarks used for localization~\cite{schmuck2019ccm,campos2021orb}. Collaborative dense systems obtain that geometry from depth sensors~\cite{yugay2025magic,xu2025mac,hu2023cp,deng2025mne,chen2026coma}, which reduce cross-agent alignment to a rigid transform but add payload, power, and calibration cost. A monocular RGB camera is a lightweight alternative, especially for small aerial platforms.

Monocular collaboration is harder than its RGB-D counterpart. Each robot starts in an independent coordinate frame and reconstructs geometry at an uncertain scale~\cite{campos2021orb}, so a shared region must supply enough evidence for a relative similarity transform rather than only a rigid displacement~\cite{lipson2024multi,zhou2026multi,li2025mang,cao2026mags}. Outdoor scenes compound this: viewpoints differ substantially, overlap is narrow, and repetitive facades or weakly textured roads make sparse feature association ambiguous. An incorrect inter-agent link distorts an entire previously consistent local map, while a missed link leaves otherwise useful observations in disconnected maps.

Learned feed-forward reconstruction models such as DUSt3R~\cite{wang2024dust3r}, MASt3R~\cite{leroy2024grounding}, and VGGT~\cite{wang2025vggt} regress dense pointmaps and pixel-level correspondences directly from RGB images, and already drive dense monocular SLAM~\cite{murai2025mast3r,maggio2026vggt,duisterhof2025mast3r}. Those systems assume a single continuous trajectory evolving under one gauge. We observe that the priors themselves supply what the multi-agent case is missing: their pointmaps anchor scale across independently reconstructed maps, and their dense correspondences permit geometric verification over a whole image region. 

\begin{figure}[t]
  \centering
  \includegraphics[width=0.9\columnwidth]{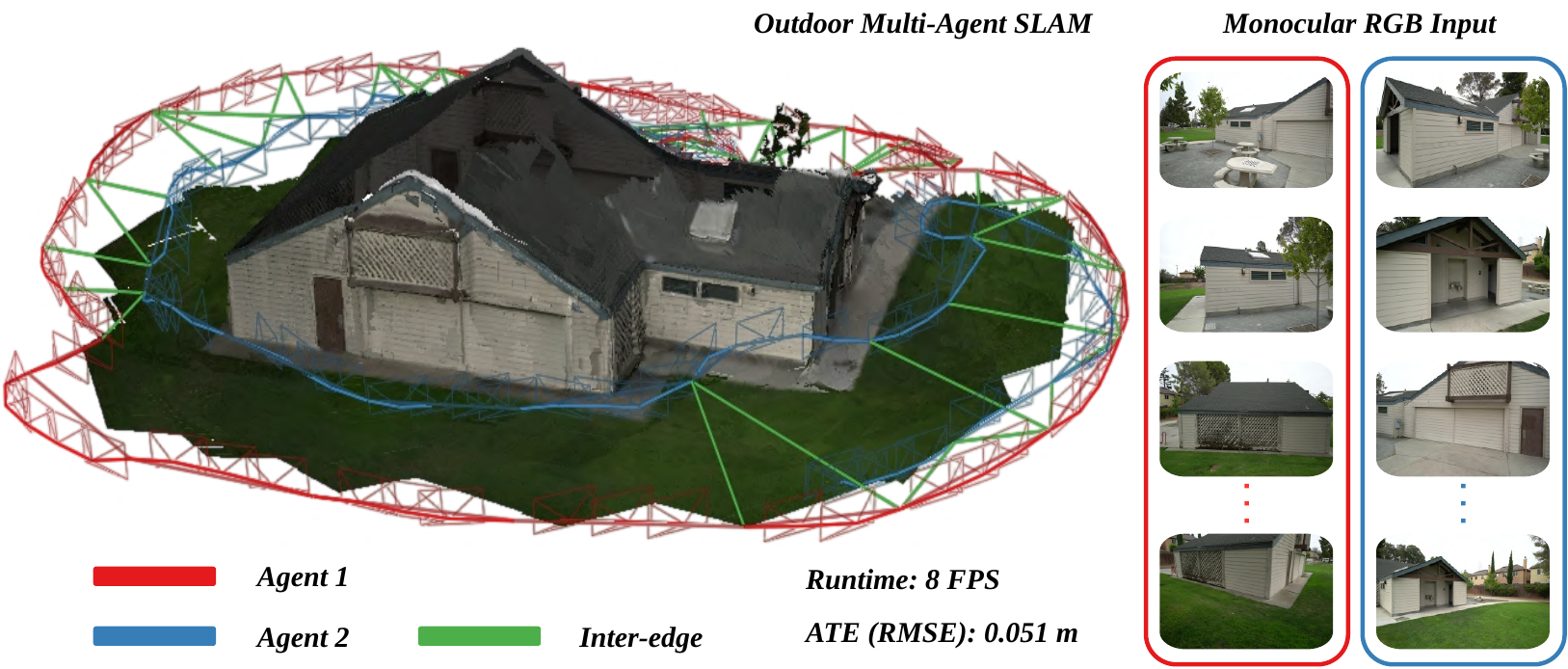}
  \caption{\ours{} combines two monocular RGB streams into a shared dense reconstruction of Barn. Red and blue trajectories belong to different agents; green links are verified inter-agent constraints. The reported ATE is 0.051\,m with 8 FPS. The code and project website is available at \textcolor{magenta}{\texttt{\href{https://como3r-slam.github.io}{https://como3r-slam.github.io}}}.}
  \label{fig:teaser}
  \vspace{-1.2\baselineskip}
\end{figure}

\begin{figure*}[t]
  \centering
  \includegraphics[width=0.99\textwidth]{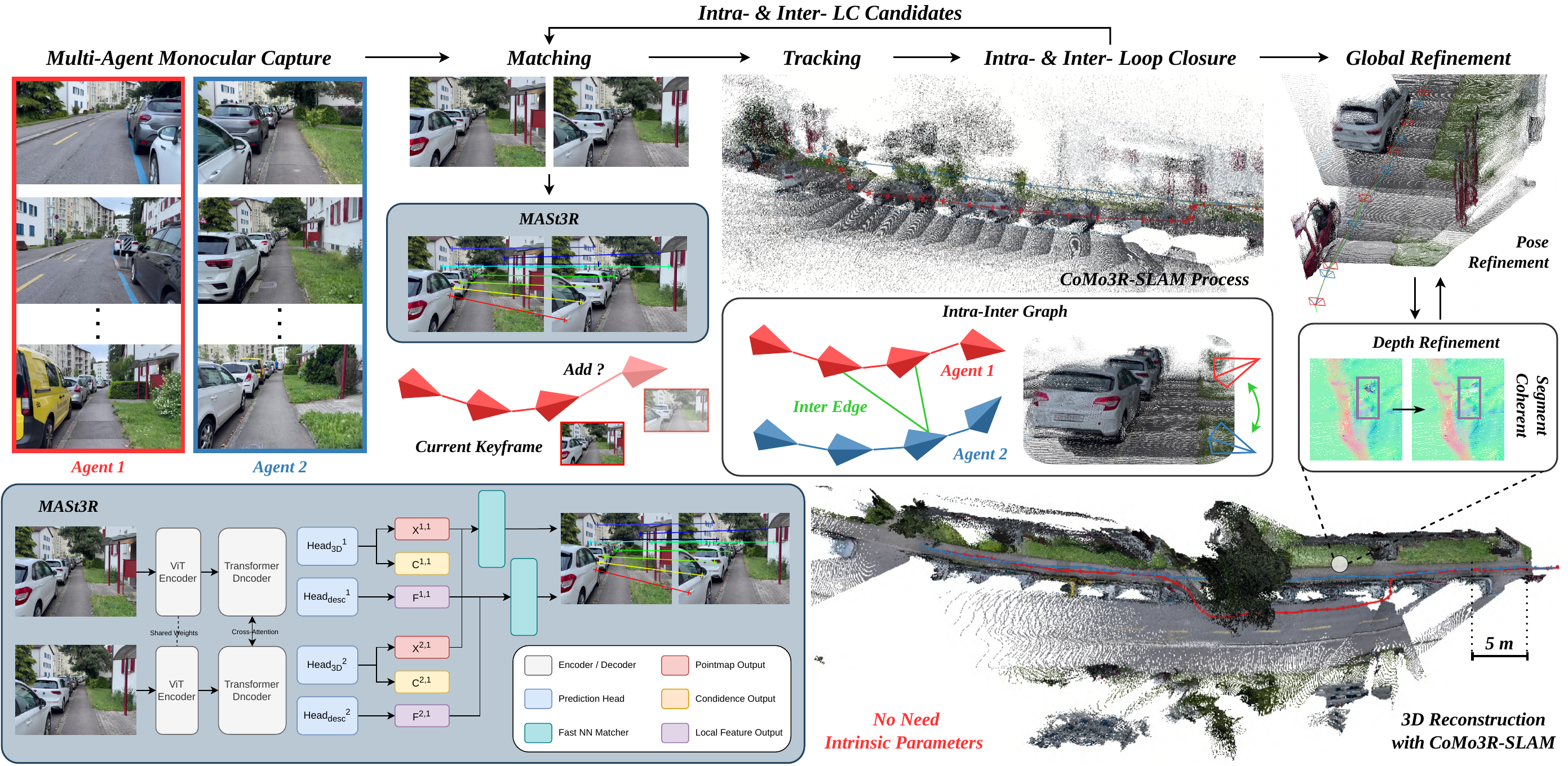}
  \vspace{-0.5\baselineskip}
  \caption{System overview. Independent monocular front-ends use the MASt3R prior for pointmap matching, tracking, and local fusion. The coordinator retrieves and geometrically verifies cross-agent keyframes, synchronizes their map components, and refines a shared similarity graph. Terminal refinement alternates global poses and geometry-aware segment depths.}
  \label{fig:method}
  \vspace{-1.\baselineskip}
\end{figure*}

We introduce \ours{}, which places learning priors at the center of a outdoor multi-agent system. Each agent runs its own prior-guided front-end, tracking and fusing keyframes from a single RGB stream. A coordinator retrieves cross-agent keyframes over the prior's encoder features, rejects candidates without bidirectional dense geometric support, transports connected map components into a common $\simthree$ gauge~\cite{umeyama1991least}, and refines intra- and inter-agent constraints in one unified graph. A terminal alternation of pose and segment-level depth refinement~\cite{mazur2024superprimitive} lets cross-agent evidence update trajectories and surfaces together. Our contributions are threefold. First, we develop a prior-guided cross-agent association and gauge-synchronization pipeline for outdoor monocular mapping. Second, we integrate inter-agent constraints into a unified $\simthree$ graph with terminal segment-level structure refinement and CUDA-supported numerical routines. Third, we evaluate trajectory accuracy, dense reconstruction, and processing cost, including a long-horizon traversal, independently captured streams, and larger teams. Together, these components extend 3D reconstruction priors from single-agent SLAM to robust cross-agent association and scale-consistent optimization, enabling monocular outdoor dense SLAM under scale ambiguity.

\section{Related Work}
\subsection{Multi-Agent Collaborative Visual SLAM}
Classical collaborative SLAM~\cite{schmuck2019ccm,schmuck2021covins,tian2022kimera,lajoie2020door} follows a pipeline of per-agent tracking, place recognition, inter-agent loop closure, and pose-graph optimization, relying on sparse features for cross-agent constraints. Those associations are dependable in textured, controlled settings and degrade precisely where outdoor teams operate: narrow overlap, wide viewpoint change, and repetitive structure. Dense collaborative methods replace the sparse map with neural fields~\cite{hu2023cp,deng2025mne} or Gaussians~\cite{yugay2025magic,xu2025mac,chen2026coma} and fuse per-agent submaps into renderable geometry, but nearly all of them read RGB-D, whose depth channel fixes each agent's scale in advance and reduces cross-agent alignment to a rigid transform. Monocular collaboration meanwhile remains sparse~\cite{lipson2024multi}, indoor~\cite{li2025mang}, feature-based~\cite{cao2026mags}, or offline~\cite{zhou2026multi}. Online outdoor multi-agent dense SLAM from monocular RGB alone is therefore largely unexplored.

\subsection{Dense Monocular Mapping and Learned 3D Priors}
Feed-forward reconstruction networks open a route to both of these obstacles at once, i.e.\ matching and alignment. From a raw image pair they predict correspondence grounded in 3D geometry rather than in appearance, which is what wide baselines and repetitive facades break, together with scale-consistent pointmaps that can anchor the scale of maps reconstructed independently. Concretely, DUSt3R~\cite{wang2024dust3r} regresses dense pointmaps in a common frame without known intrinsics, MASt3R~\cite{leroy2024grounding} adds dense local descriptors for explicit matching, MASt3R-SfM~\cite{duisterhof2025mast3r} extends this to unordered collections, and VGGT~\cite{wang2025vggt} generalizes it to many views. Built on them, MASt3R-SLAM~\cite{murai2025mast3r} and VGGT-SLAM~\cite{maggio2026vggt} recover globally consistent dense poses and geometry from monocular RGB, yet both still assume a single continuous trajectory under one gauge. \ours{} instead places these priors at the center of the multi-agent problem: their scale-consistent pointmaps become the scale anchor that ties independently reconstructed maps together, and their geometry-grounded correspondences become the evidence that admits or rejects every inter-agent link.

\label{sec:related}

\section{Method}
\label{sec:method}

\subsection{Architecture and Prior-Guided Front-End}
\label{sec:frontend}
Consider $N_a$ agents, each receiving an asynchronous monocular image stream. Each front-end maintains keyframe poses and dense canonical pointmaps in its own local world frame. A separate coordinator receives keyframe payloads, manages cross-agent associations, and returns optimized world poses (Fig.~\ref{fig:method}). Agents need no synchronized capture times or camera intrinsics. Maps become connected as soon as the agents observe sufficiently static scene structure.

\textbf{Learned geometry.} Given images $I^i,I^j\in\R^{H\times W\times3}$, the MASt3R prior $\mathcal F$~\cite{leroy2024grounding} predicts, all in the camera frame of $i$, pointmaps $X^i_i,X^j_i\in\R^{H\times W\times3}$, geometric confidences $C^i_i,C^j_i$, dense $d$-dimensional descriptors $D^i_i,D^j_i\in\R^{H\times W\times d}$, and descriptor confidences $Q^i_i,Q^j_i$. We use the pretrained model without scene-specific retraining. A keyframe $k$ stores a canonical pointmap $\hat X^k_k$ with per-pixel entries $\hat X^k_{k,m}$ in its own camera frame. Geometry is expressed through the ray operator $\psi(X)=X/\|X\|$ and $\pi(X)=(\psi(X),\|X\|)$, so a pixel observation is a unit ray together with a scalar range. Following MASt3R-SLAM~\cite{murai2025mast3r}, projective matching of a frame $f$ against $k$ first minimizes angular discrepancy on the predicted ray image, $p^\star=\arg\min_p\|\psi([X^f_f]_p)-\psi(x)\|^2$ for a queried point $x\in X^k_f$, and then refines correspondences in a local descriptor window. Invalid or geometrically inconsistent matches are removed, leaving a dense set $\mathcal M_{kf}$ whose matches carry the endpoint confidence $q_{mn}=\sqrt{Q^k_{k,m}Q^f_{f,n}}$.

\textbf{Ray-range tracking.} A pose $T_{WC_i}=\left[\begin{smallmatrix}s_iR_i&t_i\\\mathbf0&1\end{smallmatrix}\right]\in\simthree$, with $R_i\in\mathrm{SO}(3)$, $t_i\in\R^3$ and $s_i>0$, maps camera coordinates into the current world frame, so $T_{ij}=T_{WC_i}^{-1}T_{WC_j}$ maps camera $j$ to camera $i$. For a match $(m,n)\in\mathcal M_{ij}$, the residual is
\begin{equation}
 e_{mn}(T)=
 \begin{bmatrix}
 \psi(TX^j_{j,n})-\psi(\hat X^i_{i,m})\\
 \|TX^j_{j,n}\|-\|\hat X^i_{i,m}\|
 \end{bmatrix}\in\R^4.
 \label{eq:residual}
\end{equation}
The tracker estimates $T_{kf}$ by minimizing a component-wise Huber cost on the whitened residual, $E_{\mathrm{track}}(T_{kf})=\sum_{(m,n)\in\mathcal M_{kf}}\sum_{c=1}^{4}\rho([\Sigma_{mn}^{-1/2}e_{mn}(T_{kf})]_c)$, where $\rho$ is the Huber kernel of breakpoint $\kappa=1.345$ and $\Sigma_{mn}=\Sigma/q_{mn}$ with $\Sigma=\operatorname{diag}(\sigma_r^2\mathbf I_3,\sigma_d^2)$. This separates the angular noise $\sigma_r$ of a ray from the metric noise $\sigma_d$ of a range and lets the prior's own uncertainty modulate each match. Because the kernel acts per component, every entry carries its own IRLS weight $w(r_c)=\min(1,\kappa/|r_c|)$, so ray and range outliers are downweighted independently. The three ray coordinates encode two directional degrees of freedom; the range term constrains scale through the predicted geometry. A central-camera assumption is required for each frame, but different agents need not share a camera center or camera model. This formulation avoids an explicit intrinsic matrix entirely.

A keyframe is inserted when the valid or unique-match fraction falls below a certain threshold. Registered point predictions update the canonical map through a confidence-weighted running average,
\begin{equation}
 \hat X^k_k\leftarrow\frac{\hat C^k_k\hat X^k_k+C^k_f(T_{kf}X^k_f)}{\hat C^k_k+C^k_f},\quad
 \hat C^k_k\leftarrow\hat C^k_k+C^k_f,
 \label{eq:fusion}
\end{equation}
which suppresses the noise of any single short-baseline prediction. Sequential constraints and locally retrieved loop closures support each agent's local graph, which is refined by a local LM-IRLS solve. Local loop closures are handled within the agent; the coordinator receives sequential and verified inter-agent constraints as described below.

\subsection{Cross-Agent Retrieval and Geometric Verification}
\label{sec:association}
Each incoming keyframe carries its agent identity and the prior's encoder features. The released HOW-style retrieval head~\cite{tolias2020learning} transforms these features into local descriptors, aggregated into a binary ASMK representation~\cite{tolias2013aggregate}. The coordinator maintains a shared inverted-file database tagged by agent. For a query keyframe, it retains up to $k_{\mathrm{NN}}=3$ retrieved keyframes from other agents with similarity above $5\times10^{-3}$. 
For every candidate $(i,j)$ with $a(i)\neq a(j)$, the two-view prior is decoded in both directions, giving dense correspondences with validity masks and per-match confidences $q_{mn}=\sqrt{Q^i_{i,m}Q^j_{i,n}}$ in each decoded direction. Let $\nu_i$ and $\nu_j$ denote the fractions of pixels with valid high-confidence matches in the opposing view. We admit the edge only if $\min(\nu_i,\nu_j)\geq\nu_{\min}$ with $\nu_{\min}=0.1$. This test requires support in both images, reducing acceptance of one-sided associations. Dense geometry supplies evidence beyond a retrieval score, so verification stays reliable under low overlap and wide viewpoint change. 

\subsection{Synchronizing Independent Map Components}
\label{sec:gauge}
Before their first verified connection, two agents can have different origins, orientations, and scales. Directly adding their residuals to a global solve would therefore impose a poor initialization. For a first link between disconnected components, lift matched canonical points into their respective world frames, $\mathbf x_\ell=T_{WC_i}\hat X^i_{i,m}$ and $\mathbf y_\ell=T_{WC_j}\hat X^j_{j,n}$. We estimate the similarity mapping the moving component to the reference component by
\begin{equation}
 (s^\star,R^\star,t^\star)=\arg\min_{s>0,R\in\mathrm{SO}(3),t}
 \sum_{\ell=1}^{N_c}\|\mathbf x_\ell-(sR\mathbf y_\ell+t)\|^2.
 \label{eq:align}
\end{equation}
Umeyama alignment~\cite{umeyama1991least} gives a closed-form solution. With centroids $\bar{\mathbf x},\bar{\mathbf y}$, centered points $\widetilde{\mathbf x}_\ell,\widetilde{\mathbf y}_\ell$, cross-covariance $A=\frac{1}{N_c}\sum_\ell\widetilde{\mathbf x}_\ell\widetilde{\mathbf y}_\ell^\top=U\Lambda V^\top$ and $S=\operatorname{diag}(1,1,\det(UV^\top))$,
\begin{equation}
 R^\star=USV^\top,\quad
 s^\star=\frac{\operatorname{tr}(S\Lambda)}{\frac{1}{N_c}\sum_\ell\|\widetilde{\mathbf y}_\ell\|^2},\quad
 t^\star=\bar{\mathbf x}-s^\star R^\star\bar{\mathbf y}.
 \label{eq:umeyama}
\end{equation}
The resulting $T^\star=\left[\begin{smallmatrix}s^\star R^\star&t^\star\\\mathbf0&1\end{smallmatrix}\right]\in\simthree$ is left-multiplied onto every pose of the moving component, which preserves its internal relative poses while changing its world gauge.

A component tracker prevents repeated gauge transport after components are connected. Agent 0 defines the final reference; when neither component contains it, a deterministic reference is selected until a later connection joins the anchor. At least $N_c^{\min}=64$ correspondences are required for the point-based initialization. Below this threshold, the coordinator falls back to the endpoint transform $T^\star=T_{WC_i}T_{WC_j}^{-1}$, which snaps the moving endpoint onto the anchor and transports the rest of its chain by the same similarity, so the chain stays internally consistent and the subsequent global refinement absorbs the residual misalignment.

\subsection{Unified Multi-Agent Pose Refinement}
\label{sec:global}
The coordinator stores a graph $\mathcal G=(\mathcal V,\mathcal E)$ whose vertices are the poses $\{T_{WC_i}\}$ of all $N=\sum_a N^{\mathrm{kf}}_a$ keyframes in one flat global index space, with a side map $a(i)$ recording the originating agent of vertex $i$. Its edge set is $\mathcal E=\mathcal E_{\mathrm{intra}}\cup\mathcal E_{\mathrm{inter}}$, where $\mathcal E_{\mathrm{intra}}=\{(i,j)\mid a(i)=a(j),\,\alpha(i,j)=1\}$ collects the intra-agent sequential links promoted to the global graph and $\mathcal E_{\mathrm{inter}}=\{(i,j)\mid a(i)\neq a(j),\,\beta(i,j)=1\}$ the verified cross-agent links of Sec.~\ref{sec:association}. Both $\alpha(\cdot,\cdot)$ and $\beta(\cdot,\cdot)$ are binary admission indicators. Each edge stores dense matches, validity masks, and endpoint confidences. Intra-agent loop closures remain in the local graphs. For the connected map, the first anchor keyframe is fixed to remove the global $\simthree$ gauge freedom, and the remaining $N-1$ poses minimize
\begin{equation}
 E_{\mathrm{global}}=\sum_{(i,j)\in\mathcal E}
 \sum_{(m,n)\in\mathcal M_{ij}}\sum_{c=1}^{4}
 \rho\!\left([\Sigma_{mn}^{-1/2}e_{mn}(T_{ij})]_c\right),
 \label{eq:global}
\end{equation}
with $T_{ij}=T_{WC_i}^{-1}T_{WC_j}$. This lets inter-agent observations correct existing keyframe poses and scales, rather than merely applying a fixed transform to an otherwise frozen submap.

Poses are updated by the left perturbation $T_{WC_i}\leftarrow\operatorname{Exp}(\tau_i)T_{WC_i}$ with $\tau_i=(\mathbf v,\omega,\zeta)\in\mathfrak{sim}(3)\cong\R^7$, using analytic Jacobians and tangent-space operations~\cite{teed2021tangent}. Writing $\mathbf x=TX$ and $d_x=\|\mathbf x\|$, the blocks composed by the chain rule are
\begin{equation}
 \frac{D\mathbf x}{D\tau}=\big[\,\mathbf I_3\ \ {-[\mathbf x]_\times}\ \ \mathbf x\,\big],\quad
 \frac{D\psi(\mathbf x)}{D\mathbf x}=\frac{1}{d_x}\!\left(\mathbf I_3-\frac{\mathbf x\mathbf x^\top}{d_x^2}\right),
 \label{eq:jacobian}
\end{equation}
together with $Dd_x/D\mathbf x=\mathbf x^\top/d_x$, giving a $4\times7$ block per match. The two endpoints of a relative-pose edge receive $DT_{ij}/DT_{WC_i}=-\mathrm{Ad}_{T_{WC_i}^{-1}}$ and $DT_{ij}/DT_{WC_j}=+\mathrm{Ad}_{T_{WC_i}^{-1}}$, so intra- and inter-agent edges produce the same sparsity pattern of two diagonal and one off-diagonal $7\times7$ blocks. Huber IRLS weights are recomputed during refinement. CUDA routines support matching and normal-equation assembly; sparse Cholesky solves the regularized pose system
\begin{equation}
 \big(\mathbf H+\lambda\operatorname{diag}(\mathbf H)+\epsilon_r\mathbf I\big)\tau=-\mathbf g.
 \label{eq:normal}
\end{equation}
The tracker uses LM damping. The local/global backend uses escalates $\epsilon_r=\epsilon_0\cdot100^{\,r}$ from $\epsilon_0=10^{-6}$ on factorization failure, for up to six attempts. If all fail, the update is zero. After a global update, each agent refreshes its keyframe world poses. A non-keyframe retains its relative transform $T^{\mathrm{rel}}$ to the keyframe used during tracking; its final pose is recovered as $T^{\mathrm{frame}}_{WC}=T_{WC_{k}}T^{\mathrm{rel}}$. Thus corrections propagate to the full trajectory without re-tracking every input image.

\subsection{Terminal Segment-Level Structure Refinement}
\label{sec:depth}
Pose-only optimization leaves errors in predicted geometry unchanged. After the streams terminate, we refine structure with one log-scale per compact surface segment. SLIC~\cite{achanta2012slic} clusters canonical surface normals and normalized log-range into $M_k\approx200$ segments per keyframe. Pixels with confidence below $C^{\mathrm{SLIC}}_{\min}=1.5$ and segments with fewer than 16 valid pixels are excluded. For a pixel $p$ in segment $g$, the corrected point is constrained to the frozen reference ray of the pristine snapshot $\hat X^k_{k,p,\mathrm{init}}$,
\begin{equation}
 \hat X^k_{k,p}(\eta_g)=\big\|\hat X^k_{k,p,\mathrm{init}}\big\|\exp(\eta_g)\,\psi\big(\hat X^k_{k,p,\mathrm{init}}\big),
 \label{eq:depth_parameter}
\end{equation}
leaving one unknown $\eta_g\in\R$ per segment; stacking them gives $\bm\eta\in\R^M$ with $M=\sum_kM_k$, two to three orders of magnitude smaller than the per-pixel depth space. With poses fixed, the depth update uses a 3D point residual over the same verified graph:
\begin{equation}
 E_{\mathrm{data}}\!=\!\!\sum_{(i,j)\in\mathcal E}\ \sum_{(m,n)\in\mathcal M_{ij}}\sum_{c=1}^{3}
 \rho\!\left(\frac{\sqrt{q_{mn}}\big[T_{ij}\hat X^j_{j,n}(\bm\eta)\!-\!\hat X^i_{i,m}(\bm\eta)\big]_c}
 {\sigma_{\mathrm{point}}}\right)\!.
 \label{eq:depth_data}
\end{equation}
This is a separate structural objective from the ray-range pose cost, and it lets cross-agent correspondences constrain structure directly through the per-segment scalars.

For the unordered $4$-adjacent segment pairs $(g,h)\in\mathcal B$ of every keyframe, let $\Delta_{gh}=\overline{\log d^{\mathrm{init}}_g}-\overline{\log d^{\mathrm{init}}_h}$ be the reference log-range difference averaged at their shared boundary, and let $E_{\mathrm{bnd}}=\sum_{(g,h)\in\mathcal B}\big((\eta_g-\eta_h)+\Delta_{gh}\big)^2$ and $E_{\mathrm{prior}}=\sum_{g=1}^{M}\eta_g^2$. The full depth objective is
\begin{equation}
 E_{\mathrm{depth}}(\bm\eta)=E_{\mathrm{data}}
 +\lambda_{\mathrm{smooth}}E_{\mathrm{bnd}}
 +\lambda_{\mathrm{prior}}E_{\mathrm{prior}}.
 \label{eq:depth_total}
\end{equation}
The boundary term encourages agreement of the corrected log-ranges $\log d_g=\log d^{\mathrm{init}}_g+\eta_g$; the prior term penalizes departures from the learned pointmap, so $\eta_g=0$ recovers the prediction. We use $\sigma_{\mathrm{point}}=0.05$ and $\lambda_{\mathrm{smooth}}=\lambda_{\mathrm{prior}}=1$. Geometry-aware grouping reduces the number of unknowns by two to three orders of magnitude relative to per-pixel refinement, while segment boundaries drawn from normals and log-range keep the parameterization aligned with 3D structure. The depth solver retains the gradient but uses a Jacobi-style approximation to the segment Hessian: for a match whose endpoints lie in segments $g$ and $h$, the chain-ruled Jacobians $J_g=-d_i\psi(\hat X^i_{i,m})$ and $J_h=d_j\,s_{ij}R_{ij}\psi(\hat X^j_{j,n})$ contribute only their diagonal blocks to $(\mathbf H_{\mathrm{data}}+\mathbf H_{\mathrm{bnd}}+\mathbf H_{\mathrm{prior}})\Delta\bm\eta=-\mathbf g$, with $\mathbf H_{\mathrm{prior}}=\lambda_{\mathrm{prior}}\mathbf I$, while the cross-segment term $J_g^\top WJ_h$ is dropped. The boundary term is Jacobi-decoupled likewise, retaining its diagonal normal-matrix contributions while discarding the off-diagonal pair; the full gradient is preserved. This reduces each iteration to independent $1\times1$ damped Newton steps. It applies up to ten iterations per call, with damping $\lambda_{\mathrm{lm}}=0.05$ and a step clip $|\Delta\eta_g|\leq\eta_{\max}=0.5$; anchor-keyframe segments remain fixed. A terminal schedule alternates these structure updates with short global pose passes, and converges within a few rounds on all sequences. 

\section{Experiments}
\label{sec:experiments}
\begin{table*}[t]
\centering
\vspace{0.1\baselineskip}
\caption{Tracking and dense reconstruction on Tanks and Temples. ATE is averaged over agents after separate similarity alignment; accuracy and completeness are medians; F-score uses a 0.10\,m threshold. Input: sensing modality, K denoting supplied intrinsics; Map: sparse keypoint or dense surface map. B/C/I/T: Barn, Caterpillar, Ignatius, Truck. $\dagger$: concatenated stream. Dash: failure; n/r: unreported. \best{Bold}: best; \second{underline}: second best.}
\vspace{-1.0\baselineskip}
\label{tab:tnt_results}
\footnotesize
\setlength{\tabcolsep}{1.7pt}
\begin{tabular}{@{}llc*{16}{r}@{}}
\toprule
& & & \multicolumn{4}{c}{ATE RMSE [m] $\downarrow$}
& \multicolumn{4}{c}{Accuracy [m] $\downarrow$}
& \multicolumn{4}{c}{Completeness [m] $\downarrow$}
& \multicolumn{4}{c}{F-score [\%] $\uparrow$}\\
\cmidrule(lr){4-7}\cmidrule(lr){8-11}\cmidrule(lr){12-15}\cmidrule(lr){16-19}
Method & Input & Map & B & C & I & T & B & C & I & T & B & C & I & T & B & C & I & T\\
\midrule
MAGiC-SLAM~\cite{yugay2025magic} & RGB-D\,+\,K & Dense & 0.519 & 0.349 & 0.356 & 0.316 & 1.260 & 0.187 & 0.204 & 0.326 & 0.963 & 0.065 & 0.055 & 0.294 & 3.6 & 43.5 & 48.2 & 12.8\\
MAC-Ego3D~\cite{xu2025mac} & RGB-D\,+\,K & Dense & \fail & 0.361 & \fail & 0.374 & \fail & 0.345 & \fail & 0.334 & \fail & 0.175 & \fail & 0.319 & \fail & 22.7 & \fail & 20.5\\
CP-SLAM~\cite{hu2023cp} & RGB-D\,+\,K & Dense & 4.972 & 5.873 & 6.178 & 6.250 & 0.459 & 0.396 & 1.120 & 0.955 & 1.093 & 0.105 & 0.803 & 2.100 & 10.5 & 20.8 & 5.4 & 0.1\\
MNE-SLAM~\cite{deng2025mne} & RGB-D\,+\,K & Dense & \second{0.152} & 0.155 & \second{0.162} & 0.190 & 0.582 & 0.309 & 0.901 & 0.276 & 0.172 & 0.141 & 0.366 & 0.099 & 15.2 & 23.2 & 9.4 & 28.5\\
MultiSlam-DiffPose~\cite{lipson2024multi} & RGB\,+\,K & Sparse & 0.221 & \second{0.148} & 0.167 & \best{0.035} & n/r & n/r & n/r & n/r & n/r & n/r & n/r & n/r & n/r & n/r & n/r & n/r\\
CCM-SLAM~\cite{schmuck2019ccm} & RGB\,+\,K & Sparse & \fail & \fail & 3.546 & 2.248 & n/r & n/r & n/r & n/r & n/r & n/r & n/r & n/r & n/r & n/r & n/r & n/r\\
MASt3R-SLAM$^\dagger$~\cite{murai2025mast3r} & RGB & Dense & 0.205 & 0.829 & 0.283 & 0.124 & \best{0.057} & \second{0.083} & \second{0.093} & \second{0.061} & \second{0.017} & \second{0.015} & \second{0.020} & \best{0.012} & \best{77.0} & \second{69.6} & \second{66.9} & \second{78.8}\\
\rowcolor{oursgray}
\ours{} (Ours) & RGB & Dense & \best{0.051} & \best{0.102} & \best{0.144} & \second{0.053} & \second{0.086} & \best{0.076} & \best{0.070} & \best{0.056} & \best{0.015} & \best{0.013} & \best{0.017} & \best{0.012} & \second{69.3} & \best{74.8} & \best{77.9} & \best{82.4}\\
\bottomrule
\end{tabular}

\end{table*}

\begin{figure*}[t]
\centering
\includegraphics[width=\textwidth]{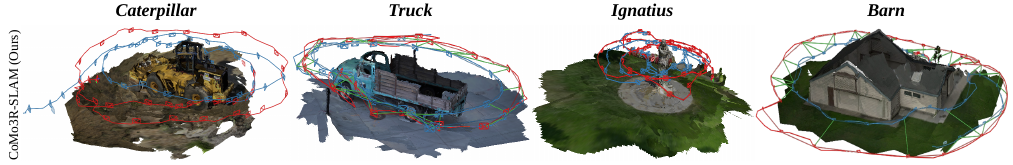}
\caption{Dense two-agent reconstructions of \ours{} on Caterpillar, Truck, Ignatius, and Barn, obtained from monocular RGB without supplied intrinsics. Red/blue: agent keyframe trajectories.}
\label{fig:tnt_recon}
\vspace{-0.5\baselineskip}
\end{figure*}

\begin{table}[t]
\centering
\caption{Tracking accuracy on Waymo (ATE RMSE [m] $\downarrow$), grouped by input modality. Supplied intrinsics, map type, markers, and the \best{bold}/\second{underline} ranking follow Table~\ref{tab:tnt_results}.}
\vspace{-0.5\baselineskip}
\label{tab:waymo}
\footnotesize
\setlength{\tabcolsep}{1.3pt}
\begin{tabular}{@{}clrrrr@{}}
\toprule
& Method & 158686 & 134763 & 153495 & 106762\\
\midrule
\multirow{4}{*}{\rotatebox[origin=c]{90}{\scriptsize RGB-D}}
 & MAGiC-SLAM & 2.409 & 4.468 & 6.113 & 15.740 \\
 & MAC-Ego3D & 8.182 & 9.255 & 11.984 & \fail \\
 & CP-SLAM & 12.277 & 10.287 & \fail & 32.641 \\
 & MNE-SLAM & 3.752 & 5.054 & 7.291 & 5.668\\
\midrule
\multirow{4}{*}{\rotatebox[origin=c]{90}{\scriptsize RGB}}
 & MultiSlam-DiffPose & \second{1.808} & 2.607 & \fail & \fail\\
 & CCM-SLAM & 2.166 & 3.805 & 22.922 & 5.436\\
 & MASt3R-SLAM$^\dagger$ & 1.937 & \second{2.035} & \best{2.914} & \best{3.172}\\
\rowcolor{oursgray}
 & \ours{} & \best{1.773} & \best{1.965} & \second{3.491} & \second{5.004}\\
\bottomrule
\end{tabular}
\vspace{-1.\baselineskip}
\end{table}

\begin{figure*}[t]
\centering
\includegraphics[width=\textwidth]{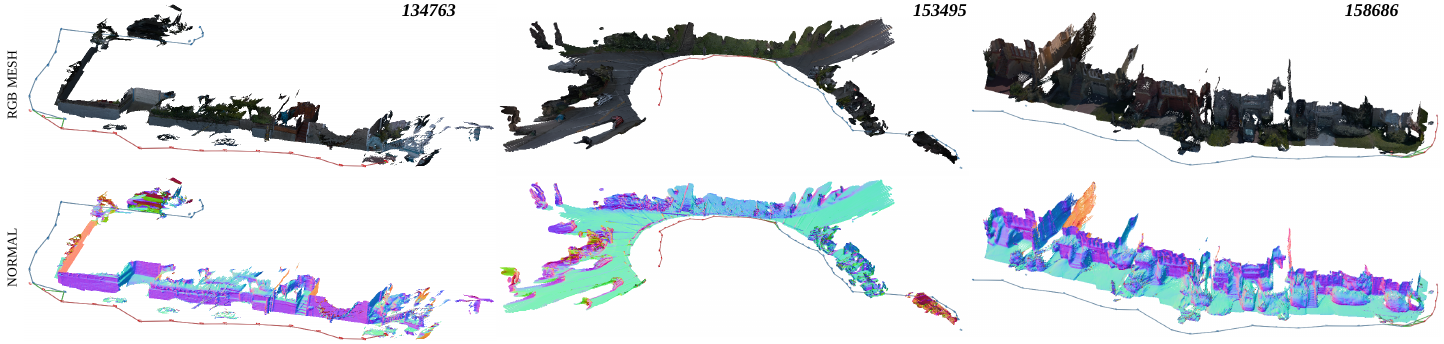}
\caption{Dense reconstructions of \ours{} on Waymo sequences 134763, 153495, and 158686 (columns), obtained from monocular RGB without supplied intrinsics. Top: fused RGB mesh; bottom: surface normals.}
\label{fig:waymo}
\vspace{-0.5\baselineskip}
\end{figure*}

\subsection{Setup and Evaluation Protocol}
We evaluate four outdoor Tanks and Temples (T\&T) scenes~\cite{knapitsch2017tanks} and four Waymo driving sequences~\cite{sun2020scalability}. Each original sequence is split into two sub-trajectories with limited overlap, provided to independent agent processes. Separately captured day and night streams with Iphone and drones examine operation beyond this split-trajectory protocol, where we sample every second frame.

We compare with four RGB-D collaborative dense systems and two RGB collaborative baselines, plus MASt3R-SLAM on concatenated streams, whose input modality and map type are listed in Table~\ref{tab:tnt_results}. MASt3R-SLAM on concatenated agent streams tests whether a single-agent prior-based system suffices, while this configuration does not preserve independent asynchronous front-ends. The main comparison uses our system without supplied intrinsics; the evaluated collaborative baselines use known calibration.

\textbf{Trajectory evaluation.} ATE is computed in metres with \textsc{evo}, using a closed-form $\simthree$ alignment to ground truth for each agent. This protocol measures trajectory shape after alignment and removes each agent's global similarity offset. Dense reconstruction and shared-map visualizations provide complementary evidence of map consistency.

\textbf{Dense evaluation.} The final confidence-filtered pointmaps are transformed into the shared frame and fused. We report the reconstruction accuracy median, completeness median, and F-score at a 0.10\,m distance threshold on T\&T. Accuracy measures reconstruction-to-reference distance; completeness reverses that direction. F-score combines the proportions of points within the threshold in both directions.

\textbf{Implementation.} The system uses PyTorch with custom CUDA routines on RTX 3080 Ti hardware. Each agent has a separate process and exchanges keyframe and pose-update messages with the coordinator. The ray and range noise settings are $\sigma_r=0.003$ and $\sigma_d=10$; the Huber breakpoint is 1.345. Online global pose refinement is triggered by an admitted inter-agent edge and uses up to 50 iterations. Final results include terminal cleanup: up to five 100-iteration pose passes, followed by at most 100 pose-depth rounds with up to ten depth iterations and 50 pose iterations per round.

\subsection{Trajectory Accuracy and Dense Reconstruction}
Table~\ref{tab:tnt_results} reports tracking accuracy and dense reconstruction on T\&T. \ours{} attains the lowest ATE on Barn, Caterpillar, and Ignatius. On Truck, MultiSlam-DiffPose is more accurate (0.035 versus 0.053\,m). Relative to concatenated MASt3R-SLAM, our errors are lower on all four T\&T scenes. The improvement is largest on Caterpillar, from 0.829 to 0.102\,m. The four-scene mean is 0.0875\,m for \ours{}, compared with 0.165\,m for MNE-SLAM and 0.385\,m for MAGiC-SLAM. Fig.~\ref{fig:tnt_recon} shows our fused two-agent reconstructions and keyframe trajectories on the four T\&T scenes. Table~\ref{tab:tnt_results} evaluates dense geometry against the T\&T reference. Both prior-based systems separate from the RGB-D collaborative baselines by a large margin in accuracy and by more than an order of magnitude in completeness: our accuracy and completeness medians average 0.072 and 0.014\,m over the four scenes, against 0.494 and 0.344\,m for MAGiC-SLAM and 0.517 and 0.195\,m for MNE-SLAM. Against concatenated MASt3R-SLAM the margin sits in completeness, where lowest on every scene, tied with the concatenated baseline on Truck, since verified inter-agent constraints fold the parts of a scene that each agent sees only partially into one surface; accuracy medians are closer on average, and Barn is the one scene where the concatenated baseline is more accurate. \ours{} achieves the highest F-score on Caterpillar, Ignatius, and Truck, with a four-scene mean of 76.1\% versus 73.1\%; the strongest gain is on Ignatius, where it exceeds the concatenated baseline by 11.0 percentage points.

Table~\ref{tab:waymo} reports tracking accuracy on Waymo, which adds forward-dominant motion, long low-parallax segments, and larger scene extent; Fig.~\ref{fig:waymo} shows our dense reconstructions and surface normals on three of the sequences. \ours{} has the lowest error overall on 158686 and 134763. However, concatenated MASt3R-SLAM is better on the latter two sequences (2.914 versus 3.491\,m and 3.172 versus 5.004\,m). CP-SLAM fails on 153495, and MAC-Ego3D fails on 106762, while MultiSlam-DiffPose fails on 153495 and 106762 in the reported runs. This demonstrates that our method serves as a valuable complement to existing approaches for monocular outdoor multi-agent dense SLAM.

The overlap sweep in Table~\ref{tab:overlap} uses $O_F\in\{5,10,20\}\%$, where $O_F$ is the fraction of keyframes with verified cross-agent matches. It is a post-verification measure, influenced by both visible overlap and association success. Increasing $O_F$ from 5\% to 10\% reduces error on every sequence, particularly Ignatius (1.628 to 0.144\,m). Increasing it further gives smaller improvements on most sequences. 

\subsection{Overlap and Component Ablations}
\label{sec:overlap}
\begin{table}[t]
\centering
\caption{Sensitivity to verified-keyframe overlap $O_F$ (ATE RMSE [m]). $O_F$ is the fraction of keyframes with verified cross-agent matches, not the per-image gate $\nu_{\min}$; $10\%$ is the default. B/C/I/T as in Table~\ref{tab:tnt_results}.}
\vspace{-0.5\baselineskip}
\label{tab:overlap}
\footnotesize
\setlength{\tabcolsep}{2.2pt}
\begin{tabular}{@{}l*{8}{r}@{}}
\toprule
& \multicolumn{4}{c}{Tanks and Temples} & \multicolumn{4}{c}{Waymo}\\
\cmidrule(lr){2-5}\cmidrule(lr){6-9}
$O_F$ & B & C & I & T & 158686 & 134763 & 153495 & 106762\\
\midrule
5\% & 0.101 & 0.229 & 1.628 & 0.081 & 2.023 & 2.358 & 4.802 & 7.045\\
10\% & 0.051 & 0.102 & 0.144 & 0.053 & 1.773 & 1.965 & 3.491 & 5.004\\
20\% & 0.050 & 0.093 & 0.076 & 0.047 & 1.819 & 1.753 & 3.134 & 4.910\\
\bottomrule
\end{tabular}
\vspace{-1.\baselineskip}
\end{table}

\begin{table}[t]
\centering
\caption{Component ablations on T\&T (ATE RMSE [m]): inter-agent loop closure (LC) and supplied calibration, with the remaining configuration fixed.}
\vspace{-0.5\baselineskip}
\label{tab:ablation}
\footnotesize
\setlength{\tabcolsep}{3.2pt}
\begin{tabular}{@{}lrrrr@{}}
\toprule
Configuration & Barn & Caterpillar & Ignatius & Truck\\
\midrule
Without inter-agent LC & 0.099 & 0.111 & 0.115 & 0.091\\
Default (with LC) & 0.051 & 0.102 & 0.144 & 0.053\\
With known intrinsics & 0.043 & 0.089 & 0.021 & 0.023\\
\bottomrule
\end{tabular}
\vspace{-0.5\baselineskip}
\end{table}

\begin{figure}[t]
\centering
\includegraphics[width=\columnwidth]{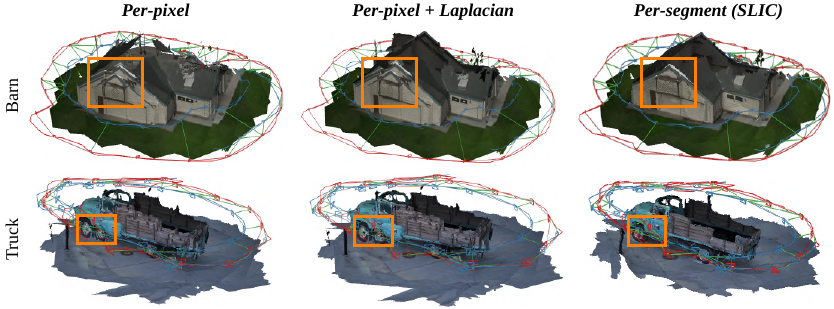}
\caption{Depth-parameterization ablation on Barn and Truck, with the remaining configuration fixed. Columns use independent per-pixel scales, per-pixel scales with Laplacian regularization, and geometry-aware segment scales. Boxes identify surface regions for comparison.}
\label{fig:depth}
\vspace{-1.2\baselineskip}
\end{figure}

\textbf{Inter-agent loop closure.} Cross-agent loop closure is what turns two independent maps into one, and removing it while retaining the rest of the pipeline raises ATE on Barn, Caterpillar, and Truck (Table~\ref{tab:ablation}). The reductions it buys are substantial, at 48.5\% on Barn, 41.8\% on Truck, and 8.1\% on Caterpillar, and the four-scene mean falls from 0.1040 to 0.0875\,m. Ignatius is the exception, worsening from 0.115 to 0.144\,m, which is consistent with ambiguity around repeated or symmetric geometry: verification can admit a constraint that improves connectivity while harming one local trajectory.

\textbf{Depth parameterization.} Segment-level refinement reduces error relative to independent per-pixel scales and to per-pixel scales with Laplacian smoothing on both tested scenes. On Barn, ATE decreases from 0.068\,m for per-pixel scales and 0.070\,m with Laplacian smoothing to 0.051\,m; on Truck, the corresponding values are 0.061, 0.058, and 0.053\,m. The gap is one of conditioning rather than smoothing: a per-pixel scale field carries far more unknowns than the verified matches constrain, and a Laplacian couples neighbours isotropically, across depth discontinuities included, so it helps only marginally on Truck and not at all on Barn. Segments cut from normals and log-range instead group pixels that share one depth degree of freedom, letting a single verified match constrain a whole patch. Fig.~\ref{fig:depth} shows the surface differences behind these numbers. 

\textbf{Calibration.} When intrinsics are supplied, we constrain pointmaps to calibrated rays and use pixel-reprojection and log-depth pose residuals \cite{murai2025mast3r}. This optional mode reduces the four-scene mean ATE from 0.0875 to 0.0440\,m, with improvements on every scene. The gain is far from uniform: Ignatius drops by 85\% and Truck by 57\%, while Barn and Caterpillar move by 16\% and 13\%. Calibration sharpens the ray basis on which the residuals act, so it pays off where the predicted rays are least reliable, which is again the near-symmetric sequence whose association is the most fragile. Accurate intrinsics therefore remain worth supplying.

\subsection{Real-world, Large-team, and Long-horizon Experiments}

\begin{figure}[t]
\centering
\includegraphics[width=\columnwidth]{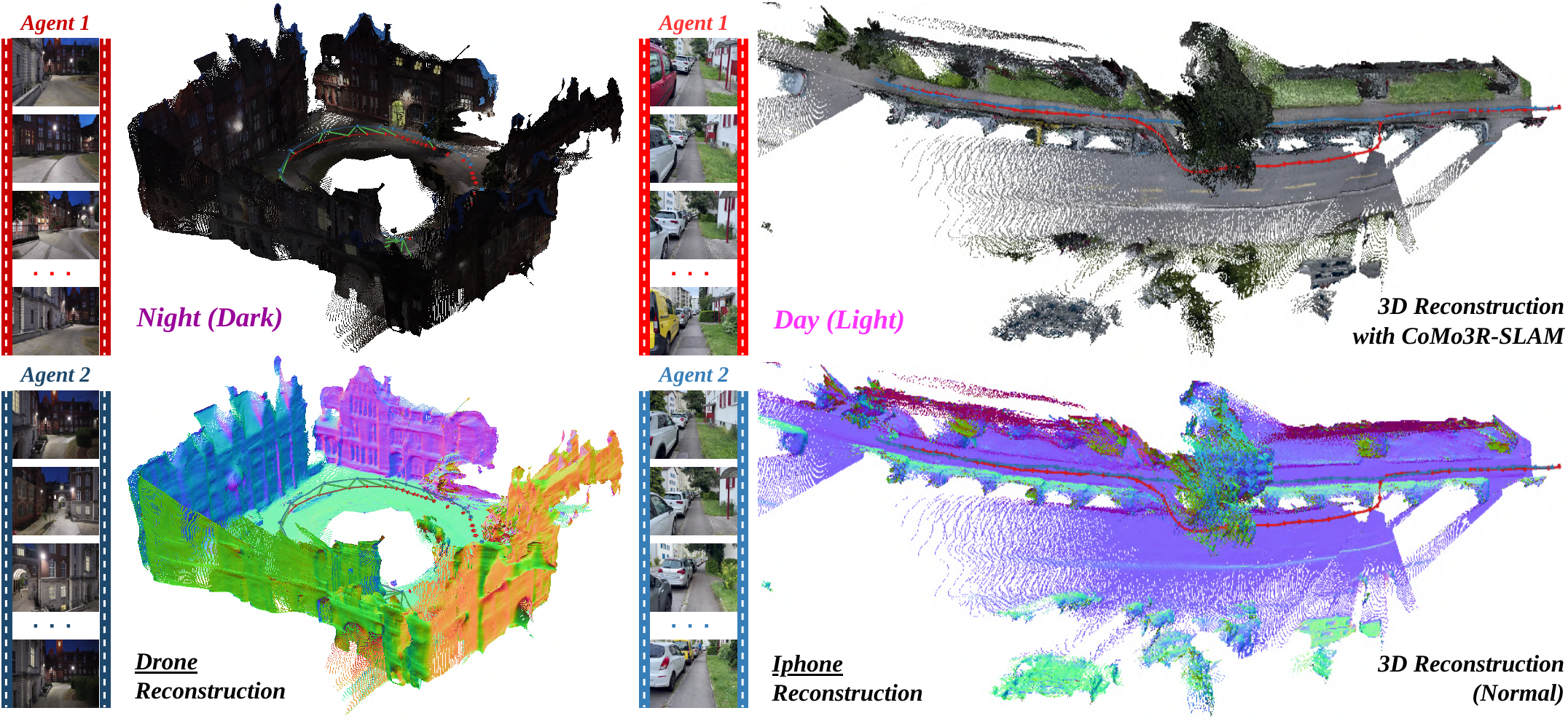}
\caption{Independently captured real-world streams without supplied camera intrinsics. Left: night-time \underline{our UAVs' online} observations and fused geometry. Right: daytime \underline{our handheld iPhone} observations and fused geometry. Each case has two agent streams; the bottom reconstructions visualize normals. Left and right are independent deployments recorded under different lighting conditions (day and night).}
\label{fig:daynight}
\end{figure}

The real-world captures in Fig.~\ref{fig:daynight} use separately recorded streams, rather than sub-sequences cut from one trajectory. The figure compares night-time UAV data and daytime handheld iPhone data, with two agents in each case. Both operate without supplied intrinsics and produce fused geometry under their respective lighting conditions. The UAV configuration streams RGB to a cloud GPU server for processing multi-agent dense reconstruction; it demonstrates remote processing with a lightweight sensor payload.

\begin{figure}[t]
\centering
\includegraphics[width=\columnwidth]{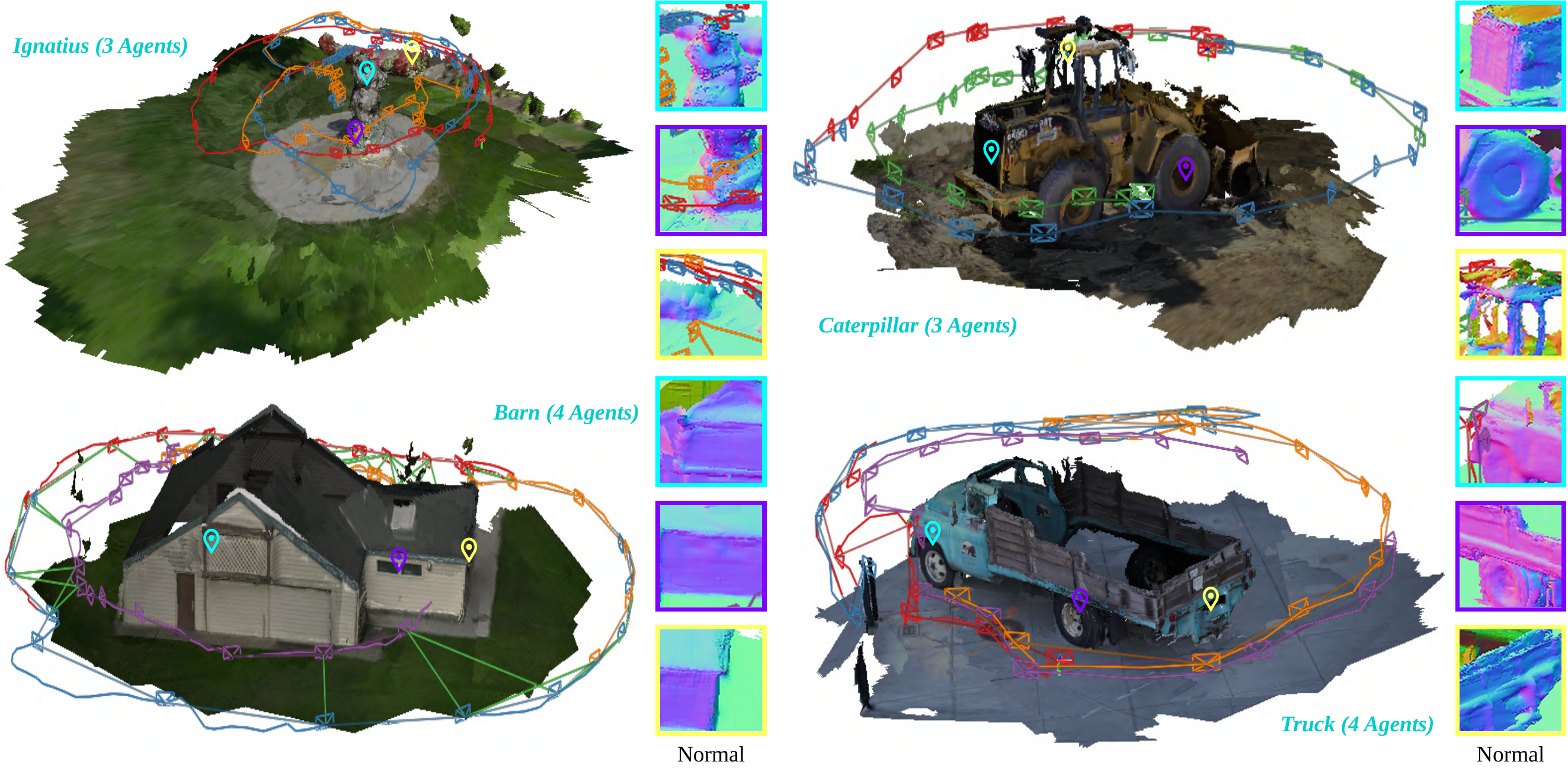}
\caption{Shared maps with different team sizes: Ignatius and Caterpillar (three agents), and Barn and Truck (four agents). Colored trajectories distinguish agents. The three- and four-agent T\&T streams are generated by partitioning a sequence into overlapping sub-trajectories.}
\label{fig:teams}
\vspace{-1.0\baselineskip}
\end{figure}

\begin{figure}[t]
\centering
\includegraphics[width=\columnwidth]{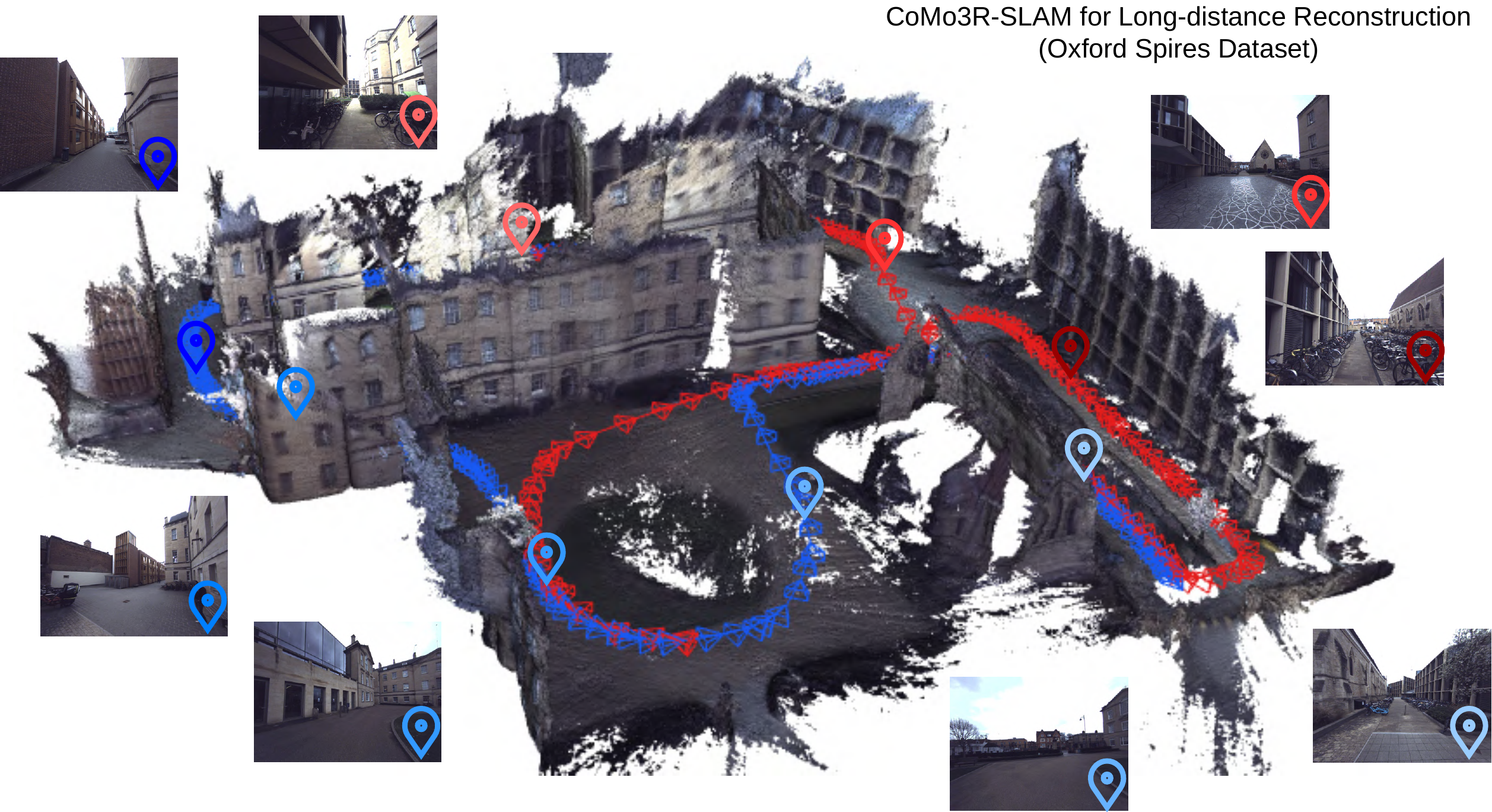}
\caption{\underline{Long-horizon} multi-agent dense reconstruction on an Oxford Spires sequence \cite{tao2026oxford}, obtained without supplied intrinsics. Red and blue frusta are the keyframes of the two agents, shown on the fused dense point cloud; the surrounding thumbnails are input views, pinned in the colour of the agent that observed them.}
\label{fig:oxford}
\end{figure}

What's more, Fig.~\ref{fig:teams} extends the shared-map visualization to three and four agents. Each additional verified connection can join previously disconnected components, after which the same global graph and depth parameterization apply.
Fig.~\ref{fig:oxford} applies the same configuration to a long-horizon outdoor sequence from the Oxford Spires dataset. Two agents traverse the college quadrangles and the connecting streets along separate routes; verified cross-agent links join their independently tracked chains into a single dense reconstruction that covers the whole site at building scale. 

\subsection{Communication, Runtime, and Memory}

\begin{figure}[t]
\centering
\includegraphics[width=\columnwidth]{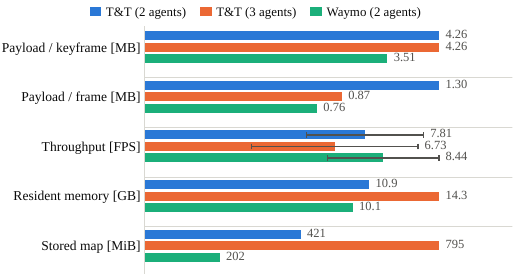}
\caption{Processing and communication costs on RTX 3080 Ti hardware. Payload is the mean message size per keyframe and per input frame; throughput is reported with $\pm1$ standard deviation; resident memory is the total over all processes. }
\label{fig:cost}
\vspace{-1.0\baselineskip}
\end{figure}

Fig.~\ref{fig:cost} reports processing, memory, and communication costs. On T\&T, moving from two to three agents reduces throughput from 7.81 to 6.73 FPS (13.8\%) and increases total RSS from 10.9 to 14.3\,GB (31.2\%). Keyframe payload stays at 4.26\,MB; the lower payload per input frame reflects a different amortized keyframe rate, rather than a smaller keyframe representation. The map grows from 421 to 795\,MiB. Waymo yields 8.44 FPS with 10.1\,GB RSS.

\section{Conclusion}
We presented \ours{}, a collaborative layer for prior-guided monocular dense SLAM in outdoor environments. Agent-aware retrieval, bidirectional geometric verification, similarity-gauge synchronization, and shared pose/structure refinement combine independently acquired RGB streams without supplied intrinsics. The evaluation shows the lowest T\&T trajectory error on three of four scenes, the highest mean F-score, competitive accuracy on driving sequences, and operation with independent captures and up to four agents.

The method depends on the learned prior, sufficient shared static structure, and successful inter-agent verification. Neither approximate learned scale nor separately aligned ATE guarantees absolute metric accuracy. Strong optical distortion and non-central cameras remain outside the demonstrated operating scope. Centralized computation, unmeasured terminal latency, and incomplete scaling measurements also limit deployment conclusions. Future work will examine shared-gauge evaluation and distributed, communication-aware mapping.

\bibliographystyle{IEEEtran}
\bibliography{main}

\end{document}